\documentclass{article}

\usepackage[numbers]{natbib}

    \usepackage[preprint]{tackling_climate_workshop_style}

\usepackage[utf8]{inputenc} 
\usepackage[T1]{fontenc}    
\usepackage{hyperref}       
\usepackage{url}            
\usepackage{booktabs}       
\usepackage{amsfonts}       
\usepackage{nicefrac}       
\usepackage{microtype}      
\usepackage{graphicx}       

\usepackage{float}          
\usepackage{subcaption}     
\title{Managing Household Waste Through Transfer Learning}

\author{%
  Suman Kunwar \\
  Faculty of Computer Science\\
  Selinus University of Sciences and Literature\\
  Ragusa, Italy \\
  \texttt{sumn2u@gmail.com} \\
}

\begin{document}

\maketitle

\begin{abstract}
As the world continues to face the challenges of climate change, it is crucial to consider the environmental impact of the technologies we use. In this study, we investigate the performance and computational carbon emissions of various transfer learning models for garbage classification. We examine the MobileNet, ResNet50, ResNet101, and EfficientNetV2S and EfficientNetV2M models. Our findings indicate that the EfficientNetV2 family achieves the highest accuracy, recall, f1-score, and IoU values. However, the EfficientNetV2M model requires more time and produces higher carbon emissions. ResNet50 outperforms ResNet110 in terms of accuracy, recall, f1-score, and IoU, but it has a larger carbon footprint. We conclude that EfficientNetV2S is the most sustainable and accurate model with 96.41\% accuracy. Our research highlights the significance of considering the ecological impact of machine learning models in garbage classification.
\end{abstract}

\section{Introduction}

The escalating global waste crisis, projected to surge by 70\% by 2050 without intervention \cite{kaza_what_2018}, demands innovative solutions. Diverse waste management techniques, from source reduction to education initiatives, strive to combat this issue \cite{mridha_intelligent_2021}. Yet, the absence of a standardized waste classification system results in regional disparities \cite{ferronato_waste_2019}, emphasizing the need for efficient waste identification, crucial for integrated solid waste management \cite{fadhullah_household_2022}. Recent advancements leverage deep learning (DL) models to streamline waste sorting and management \cite{liu_image_2022}. These models, like RWNet and ConvoWaste, exhibit high accuracy, emphasizing the role of accurate waste disposal in mitigating climate change and reducing greenhouse gas emissions. Some studies incorporate IoT and waste grid segmentation to classify and segregate waste items in real time \cite{m_technical_2023}.
 
 Integration of machine learning (ML) models with mobile devices presents a promising avenue for precise waste management \cite{kunwar_suman_2023, narayan_deepwaste:_2021}. As the advancements are happening, the computing related climate impact of ML models are not explored much. The Information and Communications Technology (ICT) industry contributes to about 1.4\% of total global greenhouse gas (GHG) emissions. Out of this percentage, roughly one-third of the emissions are due to the production and management of physical materials \cite{kaack_aligning_2022}.

Using transfer learning (TL) \cite{5288526} various waste classification techniques has been purposed \cite{poudel_classification_2022, vo_novel_2019} that shows promising results. However, the number of classes used here are quite limited and does not talks about the operational carbon emissions. Besides this, models like EfficientNetV2 that have faster training speed and better parameter efficiency has not been tested \cite{tan2021efficientnetv2}.  

The following is a summary of the contributions made by our paper:

\begin{itemize}
\item This study provides a comprehensive evaluation of transfer learning models, taking into account various settings and hyperparameters, with a particular emphasis on the Garbage dataset \cite{garbage_dataset_2023}.
\item We have employed advanced model enhancement techniques to enhance the performance of the chosen model.
\item We benchmarked the computational carbon emissions of tested models. These can serve as metrics for sustainable computing for ML models. 
\end{itemize}
The paper is structured as follows: Section 2 reviews related work. Section 3 describes the dataset used in this study and introduces the methodologies applied using the TL approaches. Section 4 presents the results and provides an analysis. Section 5 discusses the findings and concludes the paper.

\section{Related Works}
The growing challenge of waste management has spurred research into automated classification methods using deep learning. This section explores existing works related to our study, focusing on transfer learning applications, deep learning architecture comparisons, and dataset and model choices. Several studies have demonstrated the effectiveness of transfer learning for waste classification. Lilhore et al. achieved a 95.45\% accuracy for two waste categories using a hybrid CNN-LSTM model with transfer learning, highlighting its potential for efficient classification \cite{lilhore_smart_2023}. Similarly, Wulansari et al. employed transfer learning for medical waste classification with an impressive 99.40\% accuracy \cite{wulansari_convolutional_2022}, showcasing its adaptability to diverse waste types. While these studies offer valuable insights, our work specifically focuses on organic and residual waste classification, exploring the impact of transfer learning on both accuracy and training time for VGGNet-16 and ResNet-50 architectures.

Comparing the performance of different deep learning architectures is crucial for identifying optimal solutions. Mehedi et al. compared VGG16, MobileNetV2, and a baseline CNN for waste classification, with VGG16 achieving a 96.00\% accuracy \cite{mehedi_transfer_2023}. Huang et al. proposed a combination model utilizing VGG19, DenseNet169, and NASNetLarge with transfer learning, achieving 96.5\% and 94\% accuracy on two datasets \cite{huang_combination_2020}. These investigations demonstrate the effectiveness of pre-trained models and model fusion, while our work delves deeper into the performance variations between VGGNet-16 and ResNet-50 for organic and residual waste, analyzing their representation capabilities through dimensionality reduction techniques.

The choice of dataset and model architecture plays a critical role in determining the success of waste classification systems. Srivatsan et al. used pre-trained models on the CompostNet dataset for 7-class waste classification, achieving 96.42\% accuracy with DenseNet121 \cite{srivatsan_waste_2021}. Das et al. created a new 17,628-image dataset for 11 trash categories and compared ResNet152, DenseNet169, and MobileNetV2, with DenseNet169 achieving 93.10\% accuracy \cite{vasant_trash_2023}. These examples highlight the importance of dataset size and diversity, as well as the impact of model selection on specific tasks.

\section {Materials and Methods}
 
To classify the waste dataset, a TL approach was undertaken. Various state of the art transfer learning modes are used to train the model using the PyTorch framework. These model were pre-trained with ImageNet dataset and additional data augmentation techniques such as flipping, rotation, cropping, width and height were applied randomly. Additional layer such as GlobalAveragePooling2D, Dense and Dropout were added to improve classification accuracy. Both model training and testing were conducted using the Tesla T4 x2 GPUs available on Kaggle.

\subsection{Dataset}

\begin{figure}[H]
  \centering
  \includegraphics[width=0.97\textwidth, height=7cm]{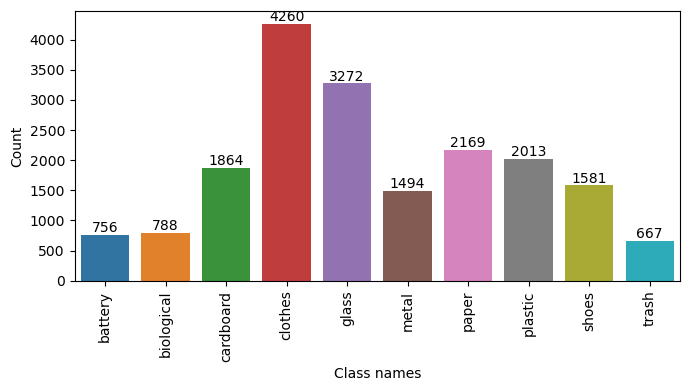}
  \caption{Image count on each class of Garbage Dataset }
  \label{fig:class_image_counts}
\end{figure}

\begin{figure}[H]
  \centering
  \includegraphics[width=0.97\textwidth, height=8cm]{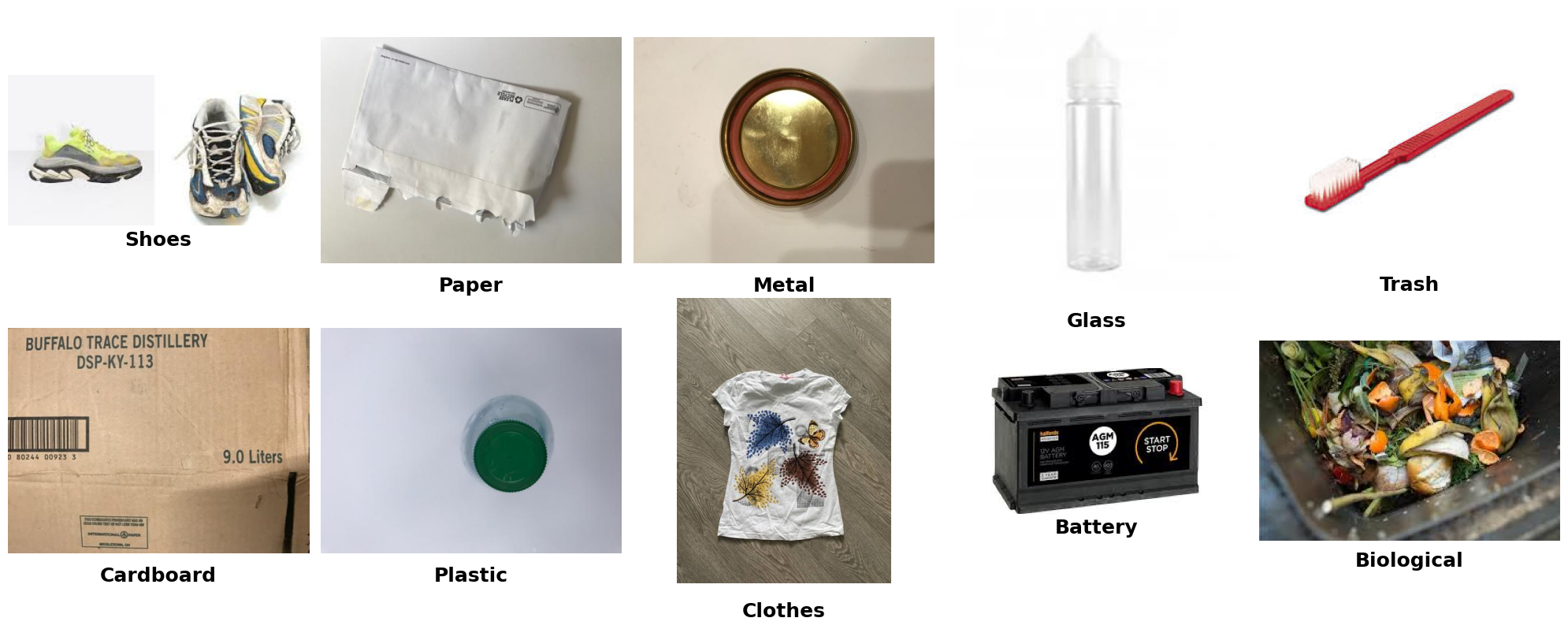}
  \caption{Sample images from each class of the Garbage Dataset.}
  \label{fig:sample_images}
\end{figure}

The waste dataset consists of 23672 images, decomposed into ten classes:  metal, glass, biological, paper, battery, trash, cardboard, shoes, clothes, and plastic.  The number of image files contain in Metal class is 1869, Glass class is 4097, Biological class is 985, Paper class is 2727,
Battery class is 945, Trash class is 834, Cardboard class is 2341, Shoes class is 1977, Clothes class is 5325 and Plastic class is 2542. These images are collected from various internet sources and also from the MWaste misclassified images. 
The dataset contains various size of images and has a class imbalance as shown in Figure \ref{fig:class_image_counts}. Some of the sample images are shown in  Figure \ref{fig:sample_images}. The class imbalance can negatively affect the training results of the model and cause it to be biased towards the largest class.

\begin{figure}[H]
  \centering
  \includegraphics[width=0.97\textwidth, height=8cm]{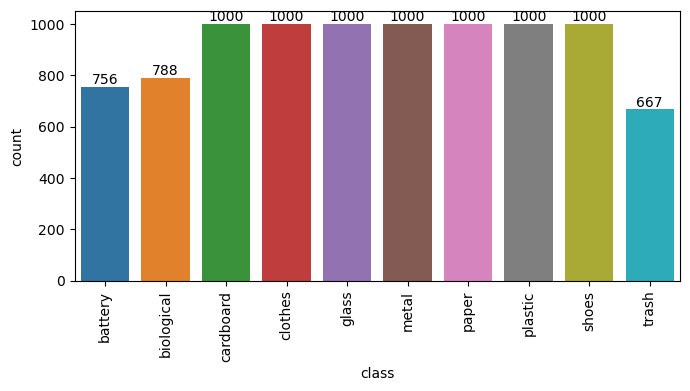}
  \caption{Sample images from each class of the Garbage Dataset.}
  \label{fig:class_imbalance_standardization}
\end{figure}

There are different approaches to solving this problem, the applicability of which depends on the problem being solved \cite{9200087}. We will use the method of insufficient sampling (random undersampling) \cite{10069499} , which consists of randomly excluding some examples from large classes. We limit the number of images in large classes to 1000 images. The updated class distribution is shown in Figure \ref{fig:class_imbalance_standardization}.

Some classes are still sparse, to solve this data augmentation techniques are applied. The average height and width are calculated from dataset and is applied to the images. The dataset is then divided into three sets: train, test and val. The train data contains 80\% of total images that will be used to train the model, the val contains 10\% and is used for checking during training and adjusting parameters whereas the test contains 10\% to evaluate the accuracy of the model on new data not used during training.  

\subsection{Training and Evaluation}

Data augmentation techniques were utilized during model training to feed augmented data into the system. Various image transformation techniques \cite{shijie_research_2017}, such as crops, horizontal flips, and vertical flips, were used for this purpose. The primary objective of this strategy was to prevent the neural network from overfitting to the training dataset, enabling it to generalize more effectively to unseen test data.

Since pre-trained models were utilized, the input dataset was normalized to match the statistics (mean and standard deviation) of those models. The model was trained under various settings, and its accuracy was subsequently evaluated. The loss was quantified using categorical cross-entropy loss. To counteract potential issues of vanishing or exploding gradients during training, which could adversely affect the parameters, the gradient clipping technique \cite{zhang_why_2020} was employed with a value set to 1.0.

\begin{figure}[H]
  \centering
  \includegraphics[width=0.99\textwidth]{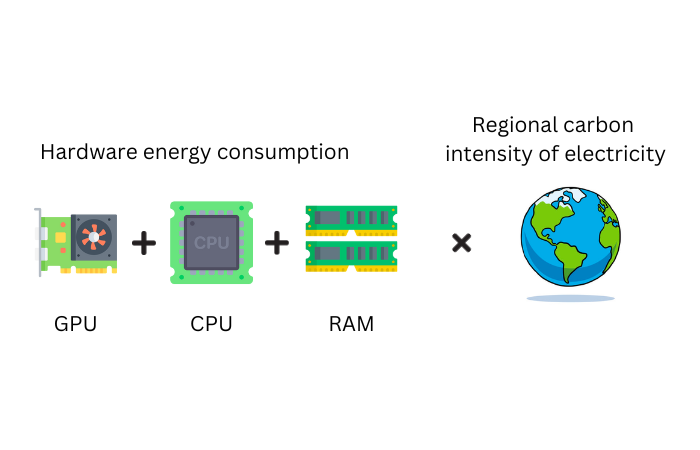}
  \caption{Codecarbon carbon emission formula}
  \label{fig:code_carbon}
\end{figure}

The Adam optimizer was used combined with multiple learning rates, due to its proven efficacy in image classification tasks \cite{10151232}. Regularization strategies, including early stopping, dropout, and weight decay \cite{8864616}, were also implemented to combat overfitting and to optimize time and resources. Various metrics such as accuracy, f1-score, recall, IoU and operational emissions were recorded, including creating and productionalizing an ML model, data storage and processing, model training, and inference with the help of codecarbon.\footnote{https://github.com/sumn2u/garbage-classification-tl}. The carbon emission from code carbon is calculated using formula shown in Figure \ref{fig:code_carbon}. 

\section{Results}
This section presents the results obtained from EfficientNetV2M, EfficientNetV2S, MobileNet, ResNet-50 and ResNet101 models trained using pre trained transfer learning with same settings. Metrics such as accuracy, recall, f1-score, IoU and the time taken to train the model are measured. Table \ref{tab:comparative_model_experimental_results} shows that the EfficientNetV2M and EfficientNetV2S has higher accuracy along with Recall, F1 score and IoU whereas ResNet50 has higher accuracy, Recall, F1 score and IoU than ResNet101. Meanwhile the MobilNet has the least value. The training time for EfficientNetV2M is highest among others and MobileNet has the lowest training time.

\begin{table}[H]
  \centering
  \caption{Comparative experimental results of EfficientNetV2M, EfficientNetV2S, MobileNet, ResNet50 and ResNet101 models with 20 epochs training time.}
  \begin{tabular}{cccccc}
    \toprule
    Model & Training time & Accuracy & Recall & F1 Score & IoU \\
    \midrule
    EfficientNetV2M   & 8035.98 sec & 96.37\% & 0.96 & 0.96 & 0.963 \\
    EfficientNetV2S  & 6016.56 sec & 96.07\% & 0.96 & 0.95 & 0.957 \\
    MobileNet   & 2163.59 sec & 68.89\% & 0.69 & 0.67 & 0.661 \\
    ResNet50  & 5868.29 sec & 94.63\% & 0.95 & 0.94 & 0.941 \\
    ResNet101   & 7371.51 sec & 94.46\% & 0.94 & 0.93 & 0.936 \\
    \bottomrule
  \end{tabular}
  \vspace{2pt} 
  \label{tab:comparative_model_experimental_results}
\end{table}

\begin{figure}[H]
  \centering
  \includegraphics[width=0.97\textwidth, height=8cm]{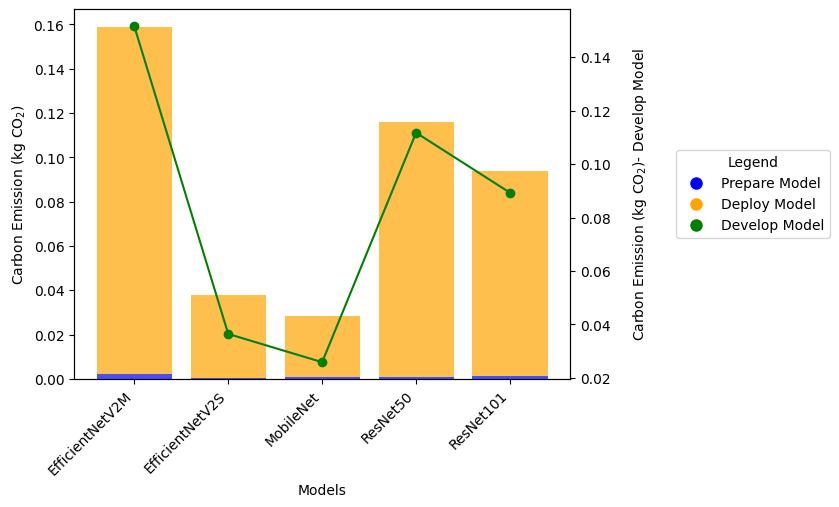}
  \caption{Carbon emission of each model at various stage}
  \label{fig:carbon_emission_barchart}
\end{figure}

The model's operational carbon emission is also measured which is shown in Table \ref{tab:carbon_emssion_results}. Here we can see that the carbon emission for EfficientNetV2S has lowest carbon emission while preparing data and moderate carbon emission on developing and deploying model. The MobileNet has lowest carbon emission while developing model and deploying model shown in Figure \ref{fig:carbon_emission_barchart}. 

\begin{table}[H]
  \centering
  \caption{Carbon emission amount at various stages of EfficientNetV2M, EfficientNetV2S, MobileNet, ResNet50, and ResNet101 models with 20 epochs training time.}
  
  \begin{tabular}{cccccc}
    \toprule
    Model & \multicolumn{3}{c}{Carbon Emission (kg CO$_2$)} \\
    \cmidrule{2-4}
     &  Prepare Data & Develop Model & Deploy Model \\
    \midrule
    EfficientNetV2M   & 0.002217 & 0.151638 & 0.156662 \\
    EfficientNetV2S  & 0.000483 & 0.036457 & 0.037580 \\
    MobileNet    & 0.000974 & 0.025888 & 0.027469 \\
    ResNet50  & 0.001111 & 0.111764 & 0.114827 \\
    ResNet101 &  0.001383 & 0.089301 & 0.092323 \\
    \bottomrule
  \end{tabular}
  \vspace{2pt} 
  \label{tab:carbon_emssion_results}
\end{table}

\begin{figure}[H]
    \vspace{3cm}
    \centering
    \begin{subfigure}[t]{0.9\textwidth}
        \centering
        \includegraphics[width=0.9\textwidth,height=8cm]{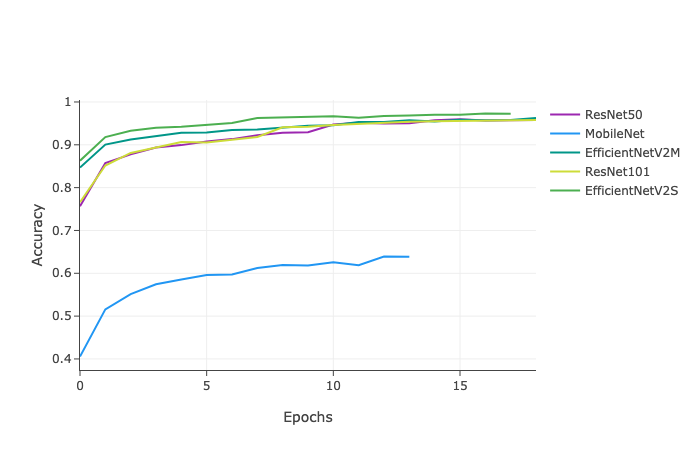}
        \caption{Training accuracy at different iterations}
    \end{subfigure}%
    \hfill
    \begin{subfigure}[t]{0.9\textwidth}
        \centering
        \includegraphics[width=0.9\textwidth,height=8cm]{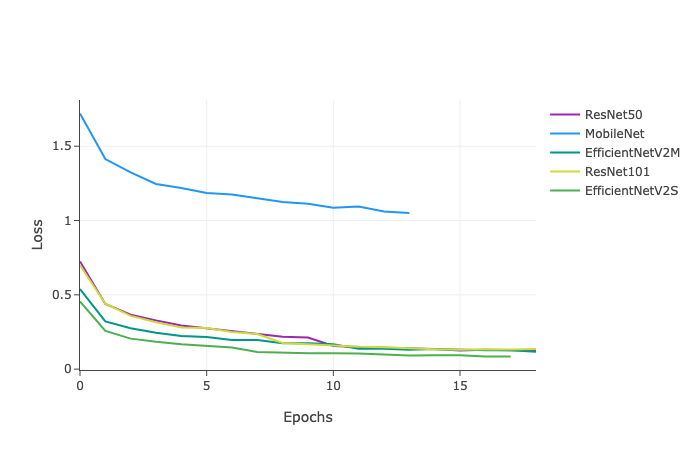}
        \caption{Evaluation loss at different iterations}
    \end{subfigure}%
    \caption{Comparison of loss and accuracy of models at different epochs}
    \label{fig:loss_accuracy_comparison}
    \hfill
\end{figure}

The accuracy of the model for 
EfficientNetV2M, EfficientNetV2S, ResNet50 and ResNet101 increase as it trains longer; whereas, the accuracy doesn't increase for MobileNet after 12 epochs shown in Figure \ref{fig:loss_accuracy_comparison} (a). Similarly the loss decreases for all models except MobileNet after 13 epochs as  depicted in Figure \ref{fig:loss_accuracy_comparison} (b).

\begin{figure}[H]
    \centering
    \begin{subfigure}[t]{0.9\textwidth}
        \centering
        \includegraphics[width=0.85\textwidth]{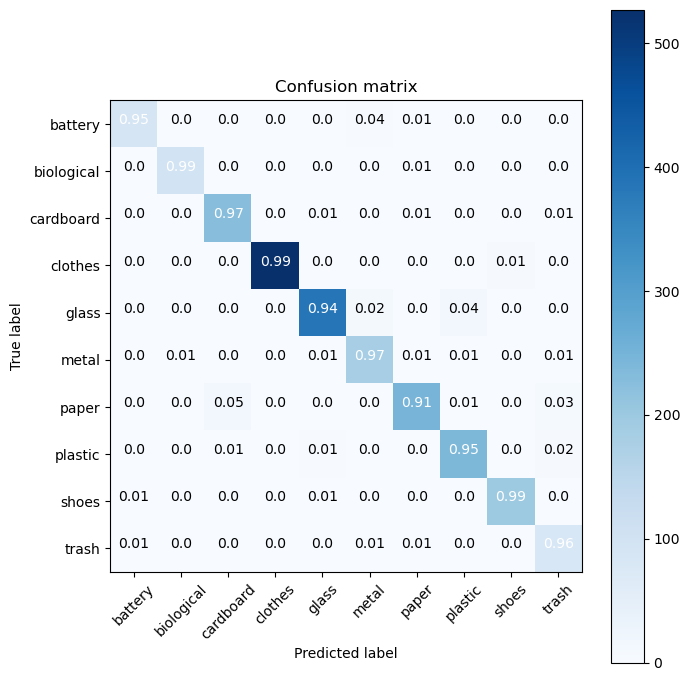}
        \caption{Confusion matrix of EfficientNetV2S model}
    \end{subfigure}%
    \hfill
    \begin{subfigure}[t]{0.85\textwidth}
        \centering
        \includegraphics[width=0.9\textwidth]{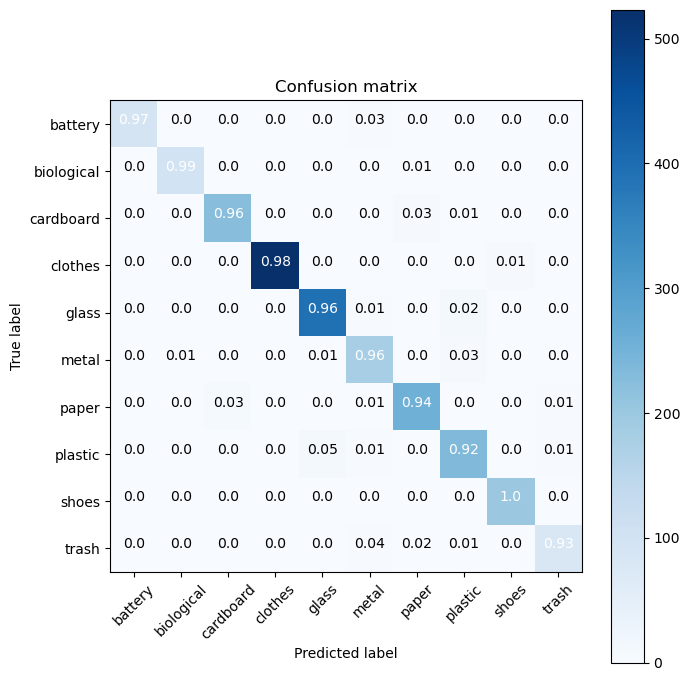}
        \caption{Confusion matrix of EfficientNetV2M model}
    \end{subfigure}%
    \caption{Confusion matrix of EfficientNetV2S and EfficientNetV2M model }
    \label{fig:confusion_matrix}
    \hfill
\end{figure}

The confusion matrix for EfficientNetV2S model is shown in Figure \ref{fig:confusion_matrix} (a) and EfficientNetV2M model in Figure \ref{fig:confusion_matrix} (b).

\begin{figure}[H]
  \centering
  \includegraphics[width=0.97\textwidth, height=7cm]{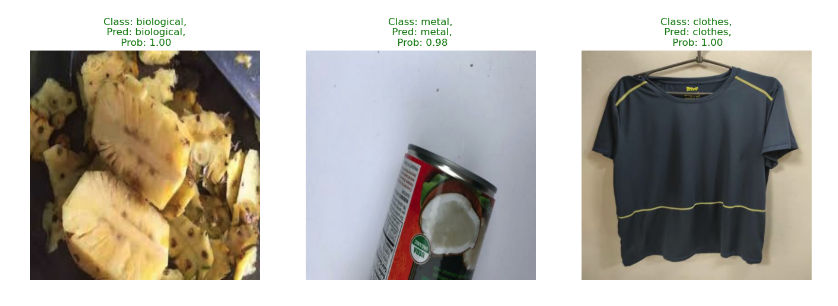}
  \caption{EfficientNetV2S model predictions on test images}
  \label{fig:model_prediction_results}
\end{figure}

From our experiment, we found that the EfficientNetV2S has higher accuracy, recall, f1-score and IoU than MobileNet, ResNet50 and ResNet101. Also, the carbon emission is moderate for preparing data, developing model and deploying model comparing to others. This model is then tested with test images as shown in Figure \ref{fig:model_prediction_results}.

\begin{figure}[H]
  \centering
  \includegraphics[width=0.99\textwidth]{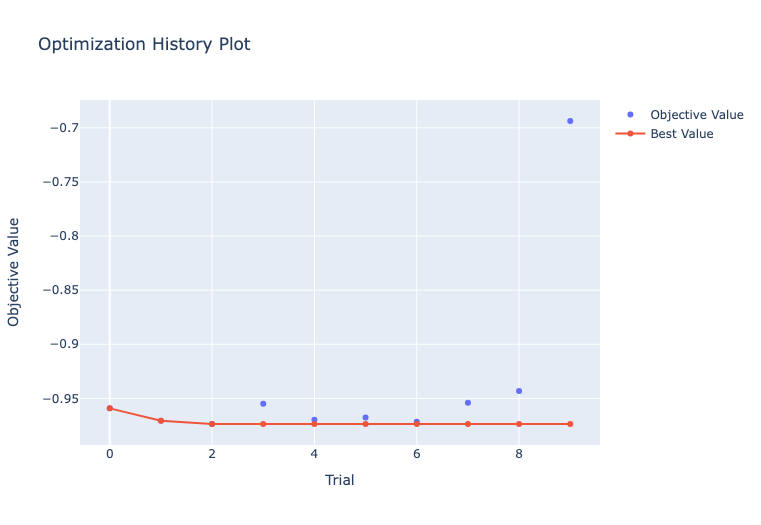}
  \caption{Optimization history plot of EfficientNetV2S model}
  \label{fig:optimization_plot}
\end{figure}

To find the best hyperparameter we used optuna, where we define the objective and get the best value that matches to our objective. The best hyperparmeter is selected and applied. Figure \ref{fig:optimization_plot} shows the optimization history plot. After applying the optimal hyper parameters the accuracy increased slightly, the new accuracy is 96.41\%.

\section{Conclusion}
During our research, we investigated various transfer learning models using a garbage dataset with pre-trained weights. We calculated the computational carbon emissions of each model and compared their accuracy, recall, f1-score, and IoU. Our findings indicate that the EfficientNetV2 family performs better than MobileNet, ResNet50, and ResNet101 in terms of accuracy, recall, f1-score, and IoU. EfficientNetV2M, which has more layers than EfficientNetV2S, produced better results but required more training time and had higher computational carbon emissions. On the other hand, the ResNet family had moderate accuracy, while the MobileNet had the least accuracy.

We discovered that ResNet50 had higher accuracy and better recall, f1-score, and IoU values than ResNet110, while requiring less training time. However, ResNet50 had a higher computational carbon emission compared to other models. After analyzing all the models' training time, accuracy, recall, f1-score, IoU, and computational carbon emissions, we concluded that EfficientNetV2S is the most sustainable model with better accuracy. Additionally, we found that adjusting the hyperparameters slightly increased the accuracy.

\bibliographystyle{unsrt}
\bibliography{tackling_climate_workshop}

\end{document}